\documentclass[sigconf]{acmart}
\usepackage{booktabs}
\usepackage{diagbox}
\usepackage{tablefootnote}
\usepackage{footnote}
\usepackage{multirow}
\usepackage{algorithm}
\usepackage{algpseudocode}

\usepackage{url}  
\usepackage{graphicx}  
\usepackage{makecell}
\usepackage{tablefootnote}
\usepackage{enumerate}
\usepackage{algpseudocode}

\usepackage{amssymb}
\usepackage{algorithmicx}
\usepackage{bm}
\usepackage{color}
\usepackage{epstopdf}
\usepackage{subfigure}
\usepackage{bm}
\usepackage{breqn}
\newtheorem{Def}{Definition}
\newtheorem{Pro}{Problem}
\newcommand{\tabincell}[2]{\begin{tabular}{@{}#1@{}}#2\end{tabular}}

\usepackage{balance} 

\AtBeginDocument{%
	\providecommand\BibTeX{{%
			\normalfont B\kern-0.5em{\scshape i\kern-0.25em b}\kern-0.8em\TeX}}}

\copyrightyear{2021}
\acmYear{2021}

\acmConference[WWW '21]{Proceedings of the Web Conference 2021}{April 19--23, 2021}{Ljubljana, Slovenia}
\acmBooktitle{Proceedings of the Web Conference 2021 (WWW '21), April 19--23, 2021,
	Ljubljana, Slovenia}
\acmPrice{}

\begin{document}
	
	\title{STUaNet: Understanding uncertainty in spatiotemporal collective human mobility}

	\author{Zhengyang Zhou}
	\affiliation{%
		\institution{University of Science and Technology of China}
		\city{Hefei}
		\country{China}}
	\email{zzy0929@mail.ustc.edu.cn}

	\author{Yang Wang}
	\authornote{Prof. Yang Wang is the corresponding author.}
	\affiliation{%
		\institution{University of Science and Technology of China}
		\city{Hefei}
		\country{China}}
	\email{angyan@ustc.edu.cn}
	
	\author{Xike Xie}
	\affiliation{%
		\institution{University of Science and Technology of China}
		\city{Hefei}
		\country{China}}
	\email{xkxie@ustc.edu.cn}
	
	\author{Lei Qiao}
	\affiliation{%
		\institution{Beijing Institute of Control Engineering}
		\city{Beijing}
		\country{China}}
	\email{fly2moon@163.com}
	
	\author{Yuantao Li}
	\affiliation{%
		\institution{University of Science and Technology of China}
		\city{Hefei}
		\country{China}}
	\email{liyuantao@mail.ustc.edu.cn}

	%
	%
	%
	%
	%
	%
	%
	
	\renewcommand{\shortauthors}{Zhengyang Zhou, Yang Wang, Xike Xie, Lei Qiao and Yuantao Li}
	
	\begin{abstract}
		The high dynamics and heterogeneous interactions in the complicated urban systems have raised the issue of uncertainty quantification in spatiotemporal human mobility, to support critical decision-makings in risk-aware web applications such as urban event prediction where fluctuations are of significant interests. 
		Given the fact that uncertainty quantifies the potential variations around prediction results, traditional learning schemes always lack  uncertainty labels, and conventional uncertainty quantification approaches mostly rely upon statistical estimations with Bayesian Neural Networks or ensemble methods. However, they have never involved any spatiotemporal evolution of uncertainties under various contexts, and also have kept suffering from the poor efficiency of statistical uncertainty estimation while training models with multiple times. To provide high-quality uncertainty quantification for spatiotemporal forecasting, we propose an uncertainty learning mechanism to simultaneously estimate internal data quality and quantify external uncertainty regarding various contextual interactions. To address the issue of lacking labels of uncertainty, we propose a hierarchical data turbulence scheme where we can actively inject controllable uncertainty for guidance, and hence provide insights to both uncertainty quantification and weak supervised learning. Finally, we re-calibrate and boost the prediction performance by devising a gated-based bridge to adaptively leverage the learned uncertainty into predictions. Extensive experiments on three real-world spatiotemporal mobility sets have corroborated the superiority of our proposed model in terms of both forecasting and uncertainty quantification.
	\end{abstract}


	

	\maketitle
	
	\section{Introduction}
	Understanding human mobility is crucial for intelligent web applications. Although many researches have delved into the predictions of human mobility including both individual trajectories and collective activities, few works have made the efforts to quantify the spatiotemporal uncertainty for human mobility. However, uncertainty quantification, which quantifies the potential variations in prediction results, is important for those applications such as epidemic forecasting, crowd management and commercial promotions where extremes are of significant interests. For instance, given the unprecedented volumes of crowds gathering in Chen Yi Square on the new year eve of 2015, the crowd monitoring system of Shanghai failed to accurately predict the abnormal variations, and hence led to a disastrous stampede which killed 36 people~\cite{zhang2017deep}. In this occasional and challenging scenario, inaccurate prediction and negligent urban management have ignored the potential uncertainty and risks which were caused by random human behaviors and complicated context influences.
	
	
	A surge of works have focused on investigating the regularities and variations in individual mobility. They explore the limits of predictability in human dynamics by measuring different types of entropy for individual trajectories~\cite{song2010limits,lu2013approaching,ikanovic2017alternative, wang2020predictability}. However, due to the inherent randomness and sparsity in individual trajectories, it is more meaningful to emphasize the uncertainty in spatiotemporal collective human mobility, which benefits location-based applications, ranging from dynamic public resource allocations and crowd-based public safety predictions to epistemic controlling. Recently, deep learning-based methods for addressing collective human mobility predictions have been widely studied~\cite{zhang2017deep, ye2019co,chen2020country}, however, all of them are incapable of capturing such uncertainties.  
	
	Regarding uncertainties, it can be further categorized into two categories, epistemic and aleatoric ~\cite{der2009aleatory,kendall2017uncertainties, kong2020sde}. Epistemic uncertainty, which can be explained with sufficient training data~\cite{postels2019sampling}, estimates the uncertainty in model parameters, while aleatoric uncertainty captures intrinsic randomness in data observations. 	
	To support uncertainty quantification in deep learning frameworks, dropout-based Bayesian Neural Networks (BNN) impose a probability distribution over learnable model parameters, and the variations on model parameters can be viewed as uncertainties~\cite{vandal2018quantifying,liu2020probabilistic,gal2016dropout,gal2017concrete,kendall2017uncertainties}. Inspired by model ensembles and physical random systems, few pioneering non-Bayesian methods such as ensemble-based~\cite{lakshminarayanan2017simple} and Brownian Motion-based methods~\cite{kong2020sde} were proposed to model the randomness of learning process in vision-related tasks. 
	Nevertheless, existing works in this aspect mostly passively learn the uncertainties from statistical estimation of testing results, which fail to internalize uncertainty extractions into the model and haven't considered the evolution of uncertainty over time and contexts. 
	
\begin{figure}[!ht] 
		\centering
		\includegraphics[scale=0.24]{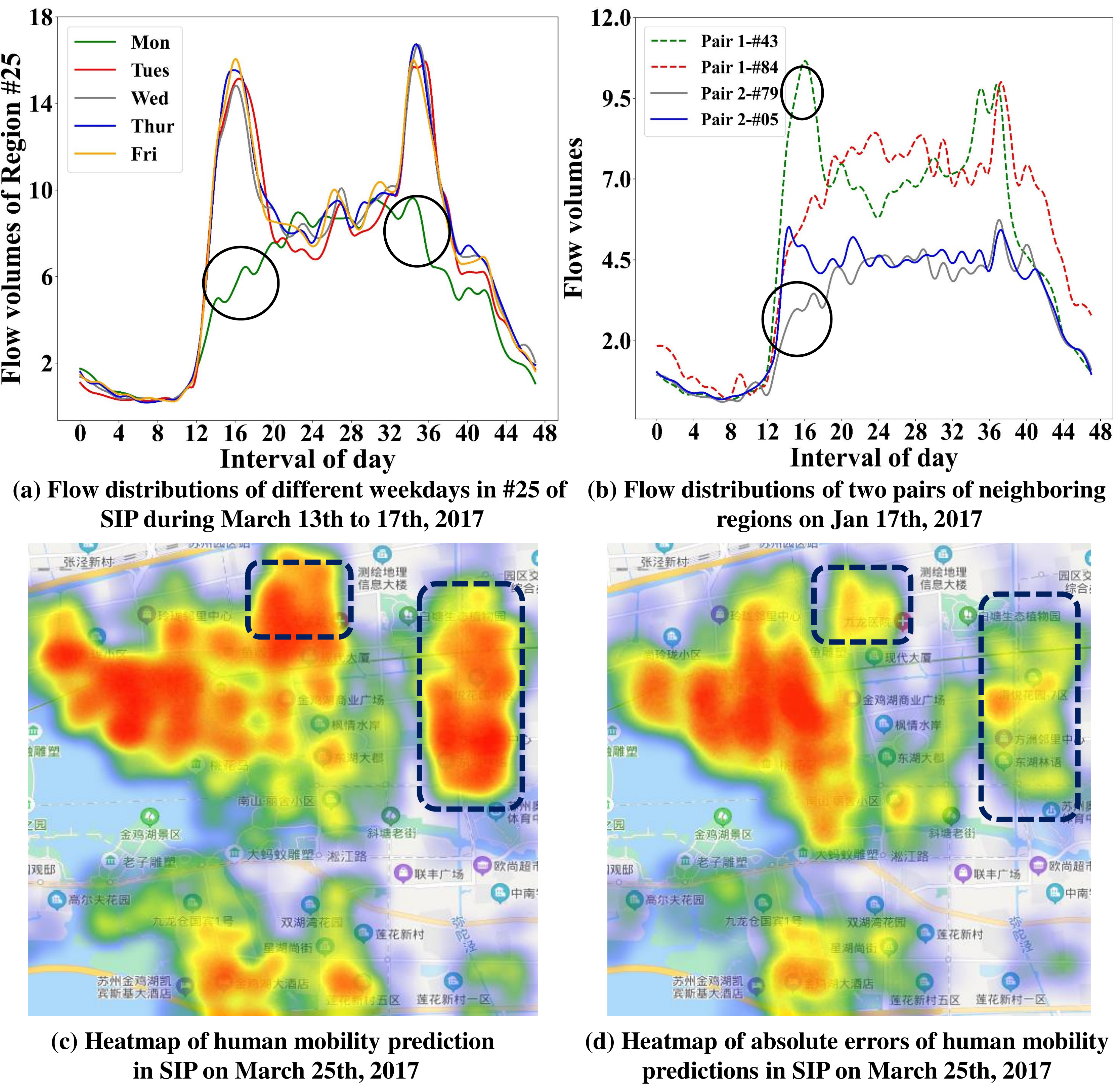}
		\caption{{Examples of uncertainty in human mobility distribution. Subfigures (a)-(b) reveal the temporal periodicities and spatial correlations in traffic flows of SIP. The circles highlight different degrees of local drifts and fluctuations of flows. Subfigures (c)-(d) illustrate the spatial heatmaps of both predictions and uncertainty quantification.}}
		\label{fig:FlowDistr}
		
	\end{figure}

	
	Differing from existing efforts, we here focus on spatiotemporal uncertainty quantification to achieve the comprehensive understanding of potential predictive fluctuations in collective human mobility, by internalizing active uncertainty extraction into our framework. Specifically, we discover that spatiotemporal tendency of human mobility can be decomposed into two parts, i.e., long-term periodicity and irregular instantaneous fluctuations. A case study of mobility distributions in Suzhou Industry Park (SIP) is illustrated  in Figure~\ref{fig:FlowDistr}(a)-(b) to reveal this observation. 
	As highlighted, the local drifts and fluctuations in different degrees may prohibit accurately  future mobility predictions, therefore there exists an urgent demand on quantifying spatiotemporal fluctuations. However, this is an intractable task because the correlations between uncertainties and spatiotemporal context factors are highly implicit and ambiguous. For instance, even the same contexts can have spatially heterogeneous influences on regions with different local functionalities, as all factors will have complex interactions with each other. Moreover, we consider the absolute error as an uncertainty indicator, and compare the spatial heatmaps of both predictions and uncertainty quantification in Figure~\ref{fig:FlowDistr}(c)-(d). From these figures, we discover an obvious task-related spatial misalignment between predictions and uncertainty quantification, and this kind of misalignment may consequently bring unexpected inefficiency to those methods which share the same feature extractors in both prediction and uncertainty learning stages~\cite{kendall2017uncertainties,lakshminarayanan2017simple,wang2019deep}.  

	In this paper, we propose a SpatioTemporal Uncertainty-aware  prediction Network (STUaNet), to address the spatial misalignment issue between uncertainty learning and mobility prediction tasks. 
	Specifically, our STUaNet consists of two modules, a spatiotemporal mobility prediction module to capture temporal dependencies with a graph-based sequential learning structure and a Content-Context Uncertainty Quantification module (C2UQ) to quantify uncertainties considering both spatiotemporal dependencies and heterogenous context influences. To actively extract  uncertainty from multi-source spatiotemporal observations in a learnable manner, we first classify the uncertainty sources of spatiotemporal data into internal content and external context, and then resolve the problem of uncertainty quantification with three main techniques of C2UQ. \textbf{(i) Neural data quality estimation.} Given the reason that data quality can significantly influence the uncertainty of predictions, we devise a similarity-based time-series quality estimation method to measure internal content consistencies and detect instaneous pattern variations in spatiotemporal data. \textbf{(ii) Context interaction learning.} To capture complicated interactions between multiple contextual factors and human mobility uncertainties, we propose a Factorization Machine-Graph Convolution Network (FM-GCN) to learn mapping functions from different contextual factors to region-level uncertainty where context-level interactions and spatial dependencies are captured by FM and GCN, respectively. \textbf{(iii) Active uncertainty learning.} To address the issue of lacking  uncertainty labels, during the training phase, we propose two weak supervised indicators and impose different spatiotemporal turbulences to imitate the Out-Of-Distribution (OOD) and random noises which correspondingly refer to epistemic and aleatoric uncertainties, and eventually enable an active weak supervised uncertainty learning. Further, we design a Gated Mobility-uncertainty Re-calibrate bridge (GMuR) to leverage the associated uncertainty, and finally boost both uncertainty learning and task-specific predictions. In summary, the main contributions of this paper are as follows. 
	\begin{itemize}
		\item To our best knowledge, we are the first to focus on the quantification of spatiotemporal uncertainty with internalized uncertainty extraction. We further innovatively re-calibrate predictions by taking advantage of uncertainty quantification, and this is an initial step on investigating how to make better use of uncertainty in optimizing predictions.
		
		\item By proposing two novel uncertainty indicators, our active hierarchical uncertainty learning enables implicit but quantifiable pseudo labels to guide unlabelled learning, and this provides prominent insights for novel indicator designs with regard to weak supervised information and active guided training schemes for consciously learning specific characteristics.
	\end{itemize}

	\section{Literature review}
	
	\textbf{Human mobility prediction.} Recent years have witnessed a surge research focusing on human mobility predictions which can be divided into two categories, the predictions of collective mobility and individual-level trajectories. Regarding the prediction of collective human mobility including traffic flows~\cite{zhang2017deep,guo2019attention} as well as taxicab pick-ups and drop-offs~\cite{ye2019co}, this issue is traditionally resolved by extracting spatiotemporal correlations with advanced variants of Convolution Neural Network (CNN) encoders. Further, by taking advantage of constructing grids into urban graph, graph-based deep learning method was introduced to forecast spatiotemporal mobility by employing full convolution blocks~\cite{yu2018spatio}. Based on~\cite{yu2018spatio}, Bai et, al. subsequently devised a graph sequence-based learning scheme to iteratively predict passenger demands~\cite{bai2019stg2seq}. 
	Regarding individual trajectory predictions, 
	given sparse and long-range trajectory sequences, \cite{liao2018predicting} and \cite{feng2020predicting} were both proposed to jointly predict the human activity and location by carefully involving sequential patterns. Besides, \cite{sun2020go} and \cite{zhao2020go} investigated individual human mobility regularities, and model personalized sequential patterns of users for next POI recommendation with Long Short-Term Memory (LSTM) network and spatiotemporal gated network, respectively. Nevertheless, none of them predict the results associated with uncertainty. 
	
	\textbf{Predictability and uncertainty in human mobility.} Previous works have been confined to the predictability and uncertainty of individual mobility based on entropy. ~\cite{song2010limits} explored the limits of predictability by studying the mobility patterns of anonymous mobile phone users, and they also identified a potential 93\% average predictability in user mobility based on  three types of entropy. These kinds of entropy respectively characterized spatial location randomness, heterogeneity of user visitation patterns and spatiotemporal order presented in personal mobility pattern. Their follow-up work~\cite{lu2013approaching} derived the maximum predictability of 88\%, by considering both stationary and non-stationary trajectories. Different from above works, ~\cite{ikanovic2017alternative} discovered that the low uncertainty in above works was highly dependent on the selected spatiotemporal scales as people didn't move in very short period. In this way, they predicted human mobility from two aspects, next location and stay time, and found an upper limit on predictability of 71\% by using natural length scale. Recently, ~\cite{chen2013uncertainty} tried to incorporate exogenous factors as uncertain factors to estimate shared mobility availability but they didn't exactly perform uncertainty quantification. Subsequently, ~\cite{geng2017partial} observed remarkable heterogeneity in individual views and further uncovered an underlying consistency between spatial and temporal human mobility in the collective spatiotemporal view, which maybe inherently related to the nature of human behavior. 
	These emerging researches only derived a time-invariant predictability and cannot be directly used to address the challenges that we are facing. However, these preliminary efforts motivated us to explore the potential variations of collective human mobility under different contexts.

	\textbf{Uncertainty quantification in deep learning. } 
	Emerging deep uncertainty quantification methods can be categorized into Bayesian and non-Bayesian lines. Bayesian methods quantify the uncertainty by imposing a probability distribution over model parameters and approximate the posterior distribution~\cite{kong2020sde}.  To infer Bayesian posterior for multi-layer perceptions, an easy-implemented variational approach was proposed by employing dropout and Monte-Carlo sampling~\cite{gal2016dropout, gal2017concrete}. Going after above works, ~\cite{kendall2017uncertainties} comprehensively analyzed how to model epistemic and aleatoric uncertainty, successfully quantified the uncertainty of image-level classification and detection based on BNN, and eventually improved the prediction reliability of risk-sensitive vision-related tasks.  On the other hand, the representative Non-Bayesian methods ~\cite{lakshminarayanan2017simple} were proposed to train multiple neural networks with different initialization and quantify uncertainty based on the statistics of prediction results. In addition, another state-of-the-art Stochastic Differential Equation (SDE) model imitated the system diffusion and captured both epistemic and aleatoric uncertainty by involving the Brownian motion~\cite{kong2020sde}. Specifically, for spatiotemporal uncertainty, ~\cite{tossebro2002uncertainty} first proposed the concept of uncertainty in spatiotemporal databases, and  then researchers have started to explore the uncertainties in large-scale climate datasets~\cite{mcdermott2019bayesian,vandal2018quantifying}, as well as uncertainty in numerical weather forecasting~\cite{wang2019deep,liu2019ensemble} with BNN methods. These works were the beginning where uncertainty quantification was introduced into spatiotemporal data mining, but they failed to capture the spatial dependency and temporal evolution of uncertainties. These above-mentioned uncertainty learning methods are also limited in addressing our spatiotemporal uncertainty quantification due to the following limitations, (i) incapability of capturing the spatiotemporal uncertainty evolution, (ii) failing to map the various context influences on uncertainties, and (iii) inefficiency of multiple times of training. 
	
	In summary, even though extensive efforts have been achieved in enhancing the prediction performances and understanding the nature of human mobility, the uncertainty issue in human mobility prediction has never been systematically considered due to  lacking awareness of spatiotemporal evolutions and context interactions. Therefore, in this paper, we will tackle this issue with a systematic perspective. 
	
	
	\section{Preliminaries and definitions}
	{\em \begin{Def}[\textbf{Urban graph and Urban regions}]
			The study area can be defined as an undirected graph, called {\em Urban Graph}. Following previous works~\cite{he2020dynamic,yao2019learning}, the whole city is discretized into a set of $N$ urban regions (e.g., POI locations, pick-up/drop-off locations and road intersections) and can be constructed as an urban graph $G(\mathcal{V},\mathcal{E})$. Here, the urban regions are composed of the vertex set $\mathcal{V} = \{ {r_1},{r_2}, \cdots ,{r_N}\}$. Given two urban regions $r_i$ and $r_j$, 	the edge ${e_{ij}} \in \mathcal{E}$ within these two urban regions can be instantiated as the geographical proximity and the potential mobility transitions.
	\end{Def}}
	
	{\em \begin{Def}[\textbf{Region human mobility intensity}]
			To describe the dynamic urban mobility intensity, we discretize time domain into equal intervals (e.g. 30 min). For region $r_i$, the human mobility intensity $H_i^t$ represents the number of active persons in region $r_i$ at the interval $t$. 
	\end{Def}}

	
	{\em \begin{Pro}[\textbf{Collective human mobility prediction with uncertainty quantification}]
			Given historical collective human mobility observations of region $i$, $H_i^1, ..., H_i^T$ where $i = 1, 2 ..., N$, we aim to simultaneously perform the point estimation of human mobility and the associated uncertainty quantification $(\widehat{H_i^{T + 1}}, \widehat{\sigma _i^{T + 1}})$ in the next interval  $T+1$, where the numerical uncertainty $\widehat{\sigma _i^{T + 1}}$ quantifies the potential variations around the prediction results. Namely, we aim to minimize the predicted uncertainty   $\widehat{\sigma _i^{T + 1}}$ while optimize tAhe prediction interval considering variations  $[\widehat{H_i^{T + 1}} - \widehat{\sigma _i^{T + 1}}, \widehat{H_i^{T + 1}} + \widehat{\sigma _i^{T + 1}}]$   to maximumly cover the ground truth.
	\end{Pro}}

	\section{Methodology}
	\subsection{Framework Overview}
\begin{figure*}[!ht] 
		\centering
		\includegraphics[scale=0.24]{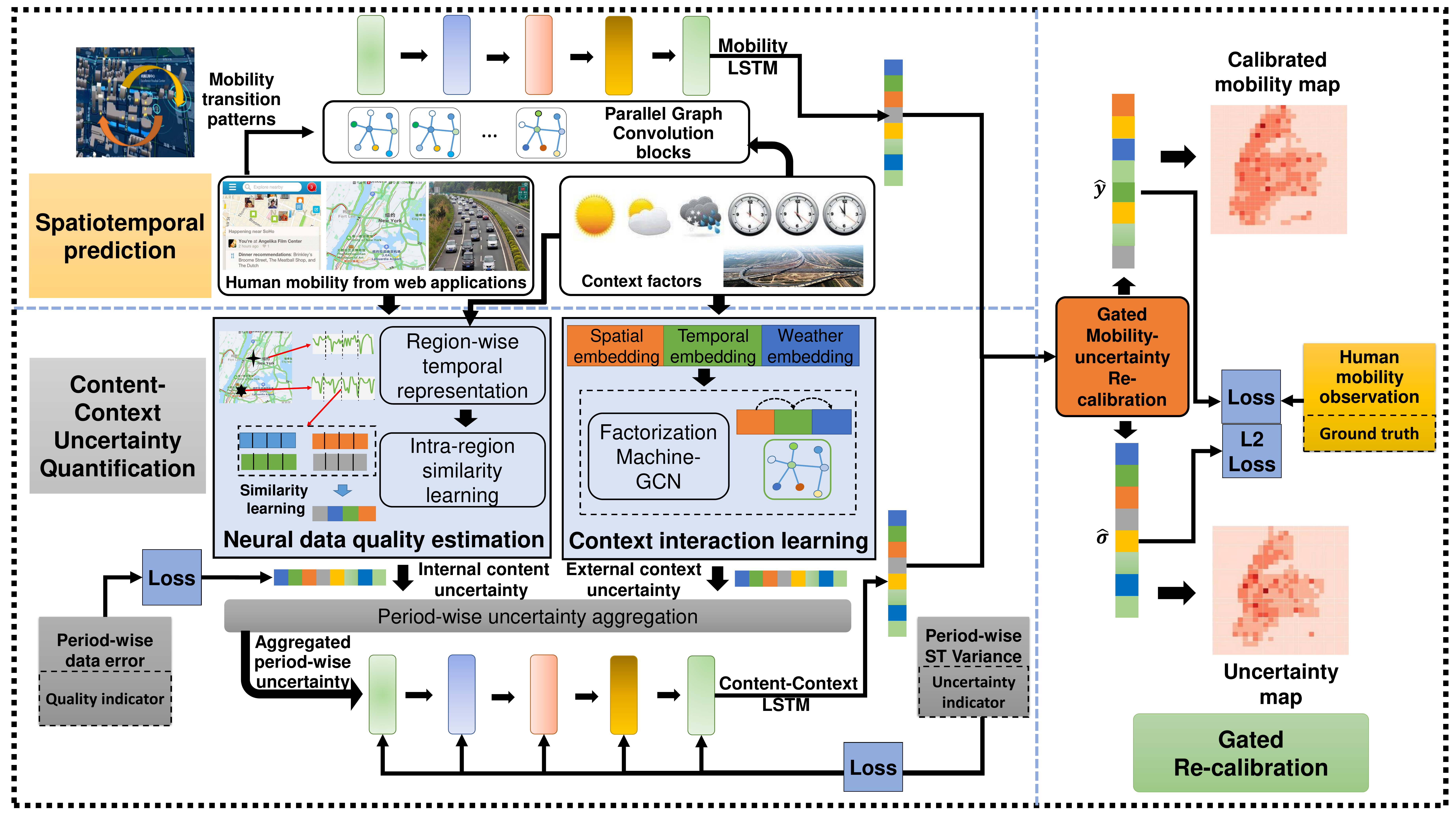}
		\caption{Framework overview of STUaNet}
		\label{fig:FO}
	\end{figure*}

	The proposed uncertainty-aware spatiotemporal prediction framework STUaNet is illustrated in Figure~\ref{fig:FO}. To disentangle  task-related spatial misalignments, we design a double-head network which consists of three components, a  spatiotemporal prediction module, a content-context uncertainty quantification module, and a gated mobility-uncertainty re-calibration bridge.  To actively learn the uncertainty, we propose a three-layer training architecture, which imitates the pure data, noisy data and OOD data by imposing different quantifiable spatiotemporal turbulences to training samples.  We will elaborate each module in the following sections.
	
	\subsection{Spatiotemporal collective human mobility prediction}
	This component serves as a spatiotemporal predictor which predicts the mobility intensity jointly in all regions in the next time interval based on historical mobility observations. As spatiotemporal forecasting  has been well investigated~\cite{bai2019stg2seq,yu2018spatio,zhang2019flow}, we here leverage a widely applied spatiotemporal model with a slight modification. We modify it by introducing the mobility transition proximity into adjacent matrix to adapt the collective mobility prediction. This predictor combines graph convolution module and an LSTM as the backbone, to extract the spatial and temporal dependencies, respectively.  
	According to the closeness and multi-level periodicity in human mobility sequences, we first construct the time period as $p$ consecutive intervals, and then retrieve a series of historical periods in the database for next-interval predictions. The selected periods are layered in three scales, i.e., the closeness period $\mathcal{P}_c$ consisting of nearest $p$ intervals, the period of daily periodicity $\mathcal{P}_d$ consisting of the same intervals as $\mathcal{P}_c$ in most recent $q$ days, and the period for summarized long-term weekly pattern $\mathcal{P}_l$ obtained by averaging the mobility intensity of the same intervals as $\mathcal{P}_c$ in each day of last week. Then we have $q+2$ groups of periods and $(q+2)*p$ intervals. To characterize the transition patterns in urban human mobility and thus better capture the dynamic spatial dependencies, we borrow the idea of cross-city migration model based on gravity systems, for their transition pattern similarities between urban mobility flows and city-wise migrations~\cite{zipf1946p}. This migration model demonstrates the fact that transitions between two specific regions are proportional to current flows in each region and inversely correlated with their spatial distances. Concisely, given time interval $t$, we instantiate the edges of urban graph as a time-sensitive mobility-involved adjacent matrix $\bm{A}^t$, where we simultaneously consider the potential transition pattern and the geographical proximity. Each element  $A_{ij}^t$ in this adjacent matrix can be formally given by, 
	\begin{equation}
	A_{i,j}^t = {e^{ - dist({r_i},{r_j})}} + \rho  \times \log (\frac{{H_i^{t} \times H_j^{t}}}{{dist({r_i},{r_j})}})
	\end{equation}
	where $dist(r_i, r_j)$ is the Euclidean distance and $H_i^{t}$, $H_j^{t}$ are the mobility intensities in $r_i$ and $ r_j$, in the corresponding interval of $t$. The scalar factor $\rho$  adjusts the proportion of region-wise transitions in overall adjacent matrix.
	
	In what follows, we respectively summarize daily periodicity $\mathcal{P}_d$ and long-term pattern $\mathcal{P}_l$ into one interval by average, yielding totally  $p+2$ intervals for sequence learning.  We employ $p+2$  GCN blocks to parallelly extract the spatial correlations and noted here we share the same adjacent matrix $\bm{A}^{t*}$ in each GCN block for the same period by averaging the adjacent matrix of all intervals in the corresponding period. To learn the temporal dependencies, we feed the feature map sequences along with the corresponding context factors (e.g. timestamps and weather)  into a mobility LSTM and finally obtain the citywide mobility intensity in the $T+1$ interval. We formulate the prediction task as,
	\begin{equation}
	\widehat{\bm{H}^{T{\rm{ + }}1}}{\rm{ = }}\bm{{\rm LSTM}}{\rm{(}}\bm{{\rm GC}}(\bm{A}^{t*},\bm{H}^{{\mathcal{P}_c}},\bm{H}^{{\mathcal{P}_d}},\bm{H}^{{\mathcal{P}_l}};\;{\bm{\theta} _{{gc}}});{\bm{\theta} _L})
	\label{eq:GCN}
	\end{equation}
	where $\bm{H}^{{\mathcal{P}_c}}$, $\bm{H}^{{\mathcal{P}_d}}$, $\bm{H}^{{\mathcal{P}_l}}$ are well-organized citywide human mobility sequences during corresponding periods. And note that $\bm{{\rm GC}}$ is the graph convolution neural network parameterized by learnable $\bm{\theta}_{gc}$, and the mobility $\bm{{\rm LSTM}}$ neural network is parameterized by learnable $\bm{\theta}_L$. 
	Here we take one of the GCN blocks to demonstrate the graph convolution \textbf{GC} by denoting the $k$-th hidden layer of GCN as $\bm{H}_G^{k}$,  
	\begin{equation}
	\bm{H}_G^k = {\rm ReLU}(\tilde {\bm{D}_A^{t*}}^{ - 1/2}{\widetilde {\bm{A}}^{t*}}\tilde {\bm{D}_A^{t*}}^{-1/2}{\bm{H}_G^{k-1}}{\bm{W}_{gc}^{k- 1})}	
	\label{eq:gcn-1}
	\end{equation}
	where ${\widetilde {\bm{A}}^{t*}} = {\bm{A}^{t*}} + {\bm{I}_N}$ and $\tilde {\bm{D}}_A^{t*}$ is the degree matrix for ${\widetilde {\bm{A}}^{t*}}$. We respectively initialize $\bm{H}_G^{0} $ as $\bm{H}^{{\mathcal{P}_c}}$, $\bm{H}^{{\mathcal{P}_d}}$ $\bm{H}^{{\mathcal{P}_l}}$ in each GCN block. $\bm{W}_{gc}^{k}$ are a series of learnable parameters that constitute the  $\bm{\theta}_{gc}$ and  we utilize ReLU as the non-linear activation function.

	\subsection{Content-Context Uncertainty Quantification}
	We propose a brand-new Content-Context Uncertainty Quantification network (C2UQ), which is tailored for spatiotemporal uncertainty learning. The C2UQ shares the same inputs with the prediction component but has a different structure for uncertainty learning. Intuitively, spatiotemporal uncertainty can arise from two scenarios, the internal data noise which suffers from two aspects,  collections and measurements, and the complicated and heterogeneous uncertainties produced by various external contexts, e.g. temporal evolution, weathers and urban events. The proposed C2UQ consists of two dedicated designed modules, neural data quality estimation and context interaction learning where these two modules are responsible for extracting uncertainties from internal data noise and external factor influences, respectively.

	\subsubsection{Neural data-quality estimation}
	Data quality estimation is a non-trivial task as it doesn't have explicit quantifiable noise labels. In our research, we find that spatiotemporal data enjoys the nice property of periodicity and closeness, and human mobility sequences usually reveal the multi-level periodicity in both long-term and short-term~\cite{feng2018deepmove}.  This intuition provides us the opportunity to detect the internal data noises and sequence pattern variations from the content perspective, by computing period-wise sequence similarities where we reuse the concept of period in Section 4.2. 
	
\begin{figure}[!ht] 
		\centering
		\includegraphics[scale=0.30]{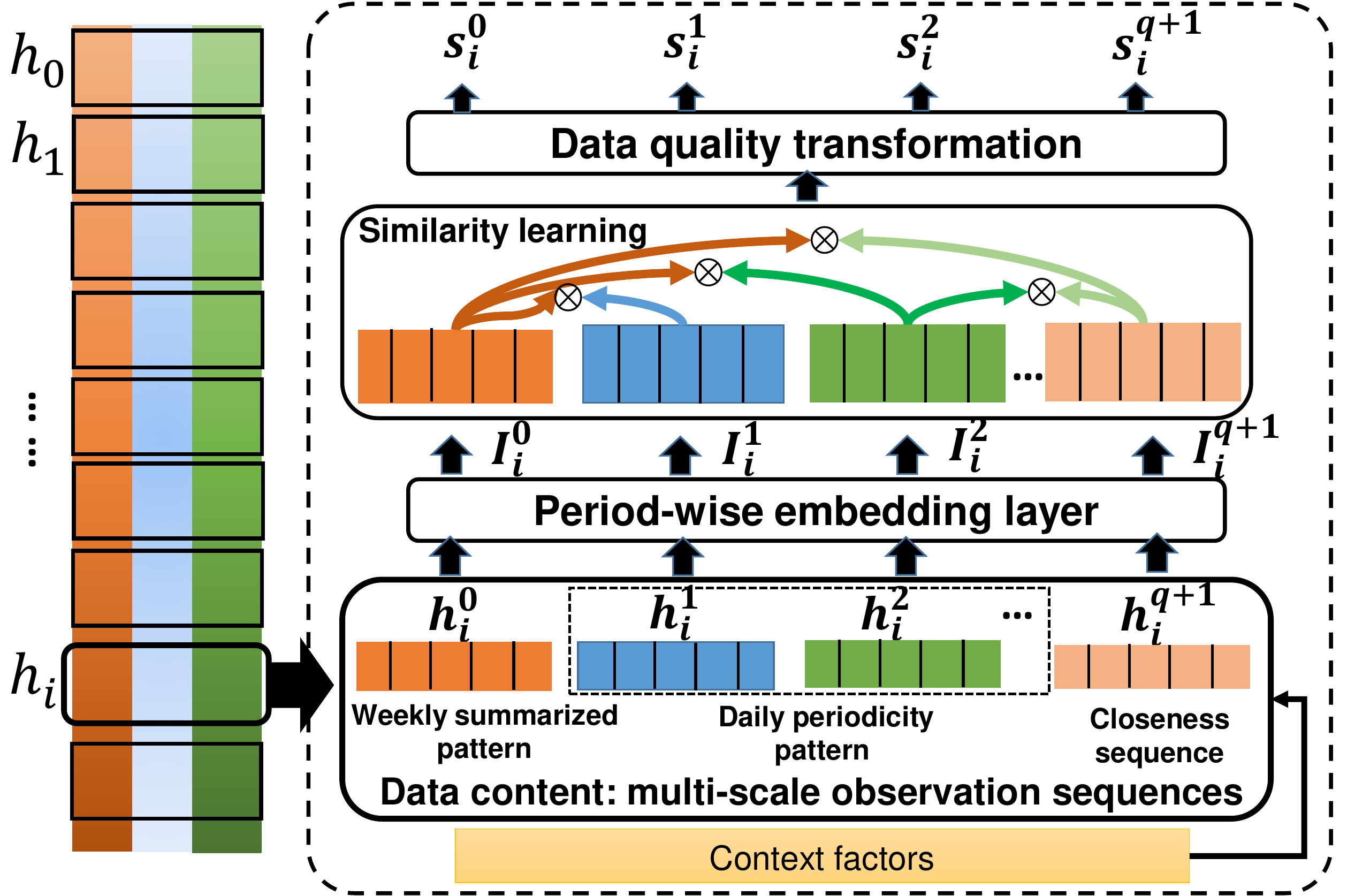}
		\caption{Structure of neural data quality estimation}
		\label{fig:neural data quality}
	\end{figure}

	To this end, we propose the multi-scale similarity-based neural data quality estimation which is illustrated in Figure~\ref{fig:neural data quality}.  The neural data quality estimation is specified to each individual region to capture the internal consistency of data observations.
	We implement this data quality estimation as follows. Firstly, we organize $q+2$ periods as a human mobility sequence for each region, i.e. closeness period $\mathcal{P}_c$, daily periodicity $\mathcal{P}_{d_i} (i\in {1, 2,..., q})$ and long-term weekly pattern $\mathcal{P}_l$. 
	For region $r_i$, we denote the selected $q+2$ historical periods of its  human mobility intensity set $\mathcal{H}_i^O$ as,
	\begin{equation}
	\mathcal{H}_i^O = \{\;\bm{h}_i^l\;, \;\bm{h}_i^{{d_1}}\;,...,\;\bm{h}_i^{{d_q}},\bm{h}_i^c,\;\} 
	\end{equation}
	and we correspondingly redefine the superscript of $\bm{h}_i^{*}$ in $\mathcal{H}_i^O$ as $\{0, 1, 2, ..., q+1 \}$ for simplicity. We then develop our model in a period-level and each period still consists of  $p$ intervals.	

	Second, we embed the context factors of respective periods and  add them into the period sequence. A neural network (e.g. fully-connected layers) is then imposed to each period sequence for learning period-level deep representations. The context encapsulated mobility intensity $\bm{I}_i^j$  during period $j$ is formulated by 
	\begin{equation}
	\bm{I}_i^j = \bm{W}_i^j  (({\bm{W}_{ex_{i}}^j}  \bm{ex}_i^j) \oplus \bm{h}_i^j) + \bm{b}_i^j
	\end{equation}
	where $\bm{ex}_i^j$ denotes the concatenated $L_c$-dimensional contextual factors of $j$-th period at the region $i$, ${\bm{W}_{ex_{i}}^j} \in {\mathbb{R}^{N \times {L_c}}}$ is the weight of  mapping function for aligning the dimensions of external factors to the same as $\bm{h}_i^j$. Here, $\bm{W}_i^j \in  {\mathbb{R}^{{L_e} \times N}}$
	and $\bm{b}_i^j\in  {\mathbb{R}^{{L_e} \times 1}}$ are  learnable weight and bias for period embedding, where $L_e$ denotes    embedded period dimension. Noted that $\oplus$ is the element-wise addition operation.
	
	Subsequently, 
	we compute the period-wise similarities 
	among all periods, and for period $m$ we have,
	\begin{equation}
	{{s}_i^m} = \frac{1}{{q + 1}}\sum\limits_{j = 1(j \ne m)}^{q + 2} {{\rm{sim}}(\bm{I}_i^m,\bm{I}_i^j)}
	\end{equation}
	where ${s}_i^m$ is the $i$-th element in vector $\bm{s}^m$, and the similarity is measured by the Hadamard product between $\bm{I}_i^m$ and  $\bm{I}_i^j$, 
	\begin{equation}
	{\rm{sim}}(\bm{I}_i^m,\bm{I}_i^j) = \bm{I}_i^m \cdot \bm{I}_i^j
	\end{equation}
	As  data noise and uncertainty of observations are inversely correlated with the period-wise similarity, we can obtain  uncertainty in  internal content view by imposing a negative exponential  function and linear transformations to $s_i^m$ where $\bm{W}_{I}^m \in \mathbb{R}^{N \times N}$, $\bm{b}_{I}^m \in \mathbb{R}^{N \times 1}$ are parameters in the learnable transformation. The citywide internal content uncertainty $\bm{U}_{I}^m$ of period $m$  is learned by,
	\begin{equation}
	\bm{U}_{\mathit{I}}^m = \bm{W}_{{\mathit{I}}}^m{e^{ - {\bm{s}^m}}} + \bm{b}_{\mathit{I}}^m	
	\end{equation}
	It is worth noting that the  weekly period $\bm{I}_i^{0}$ (corresponding to $\bm{h}_i^{l}$) is the high-level summarized mobility pattern during the same period of different days, which can be viewed as a multi-scale temporal pattern, along with  consecutive mobility intensities from $\bm{I}_i^{{1}}$ to $\bm{I}_i^{{q}}$.

	\subsubsection{Context interaction modeling}
	As pointed earlier, external contextual factors tend to interact with each other and contribute to the prediction uncertainty. For instance, mobilitIES in regions of different functionalities are with various sensitivity to extreme weather. And the mobility volumes become difficult to quantify when there exist urban events such as  concerts and accidents, because urban travelers will randomly select unhindered segments heading for their destinations. This can lead to spatially increasing mobility uncertainty. 
	With these intuitions, we propose a deep Factorization Machine-based Graph Convolutional Network (FM-GCN) to quantify context influences on uncertainty by simultaneously modeling context interactions and spatial dependencies between various external contextual factors. 
	
\begin{figure}[!ht]	
		\centering
		\includegraphics[scale=0.28]{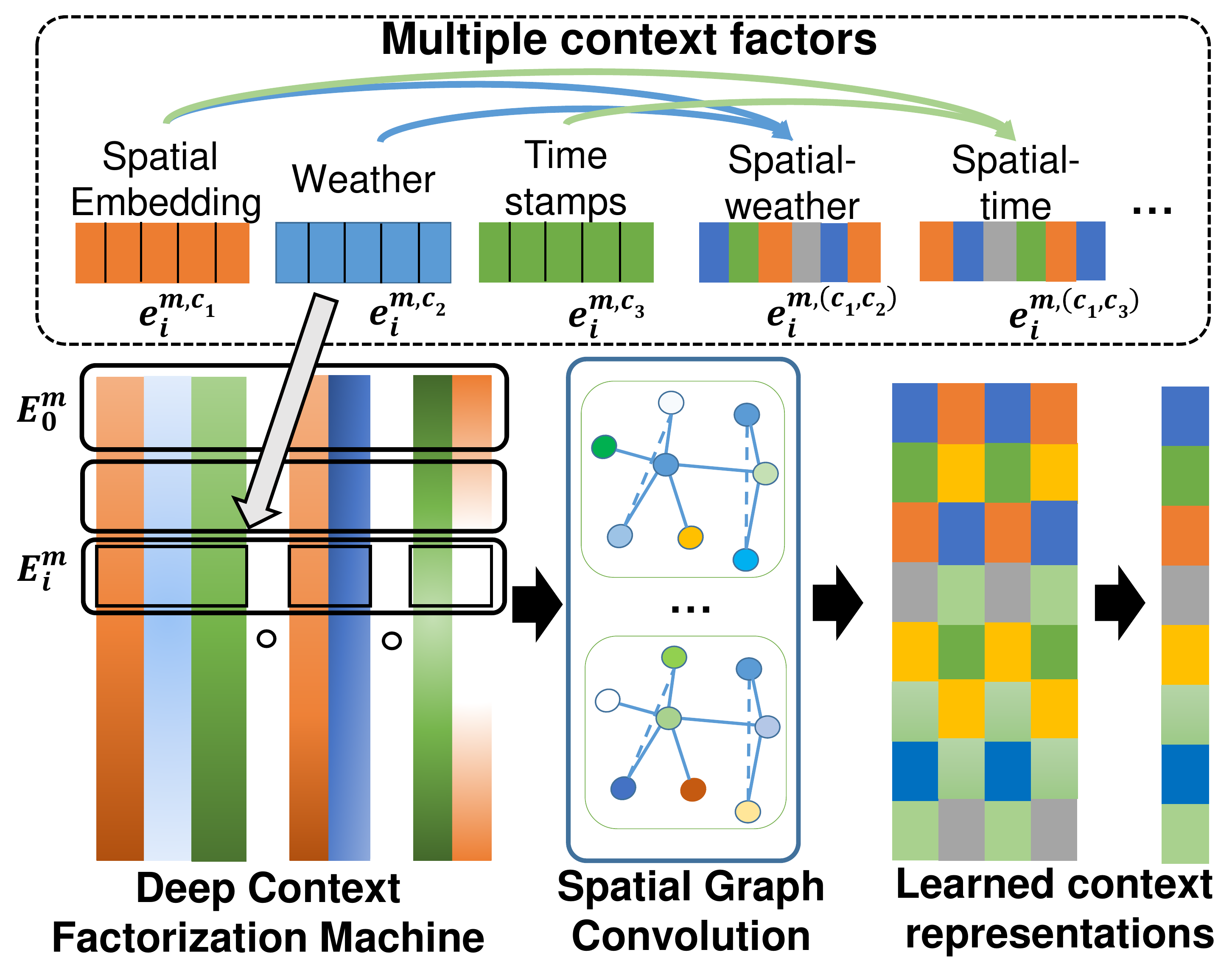}
		\caption{Details of FM-GCN}
		\label{fig:FM-GCN}
	\end{figure}

	Technically, deep factorization machine was proposed in ~\cite{lian2018xdeepfm} to extract the field-wise interactions by performing vector-level multiplication and learning  feature interactions implicitly  in recommendation system. In our work, we take advantage of this vector-level interaction modeling in FM and the spatially uncertainty propagation learning in GCN, and then seamlessly combine the FM and GCN to achieve the mappings from context interactions to spatiotemporal uncertainties. To be detailed, as Figure~\ref{fig:FM-GCN} shown, given the period $m$ and region $i$, we first embed the $Q$ categories of context factors into $Q$  vectors $\bm{e}_i^{m,c_u} (u = 1,2,...,Q)$  with the fixed-length of $L_{ce}$ as different fields in our FM, where $c_u$ represents the $u$-th category of the context factor. 
	
	Given period $m$ and region $i$, we formulate the field-wise interaction learning as, 
	\begin{equation}
	\bm{e}_i^{m,({c_u},{c_v})} = (\bm{e}_i^{m,{c_u}} \cdot \bm{e}_i^{m,{c_v}})\bm{W}_{{E_i}}^{({c_u},{c_v})}		
	\end{equation}
	where $\bm{W}_{{E_i}}^{({c_u},{c_v})} \in \mathbb{R}^{{L_{ce}}\times L_{\mathit{ie}}}$ is the learnable weight  that implements the interaction learning between $u$-th and $v$-th factors and maps the interaction embeddings into $L_{\mathit{ie}}$ dimensions. By concatenating embeddings of all fields and their counterpart interactions together, we obtain the region-wise period-level context interactions $\bm{E}_i^m \in \mathbb{R}^{ 1 \times {(Q \times {L_{ce}} + \frac{{Q(Q - 1)}}{2} \times {L_{\mathit{ie}}})}}$.

	Motivated by the  capacity of non-Euclidean spatial propagation in GCN, we hereby employ the GCN blocks to carry out the spatial influence aggregation with our mobility-involved adjacent matrix. By defining the compressed context interactions $\bm{E}_i^m$ as features of node $i$ in urban graph where $\bm{E}_i^m$ is the element in citywide context interaction tensor $\bm{E}^m$, We can perform spatial aggregation by Eq.\ref{eq:gcn-1}. The citywide uncertainty regarding external context interactions $\bm{U}_{E}^{m}$  can be learned by,
	\begin{equation}
	{\bm{U}_{E}^{m}} = \bm{{\rm GC}}(\bm{A}^m,{\bm{E}^m};{\bm{\theta} _{F_{gc}}})
	\end{equation}
	where $\bm{A}^m$ denotes the average of  adjacent matrices  at all intervals during period $m$, and $\bm{\theta} _{F_{gc}}$ is the graph convolution kernels to perform spatial context aggregation. The proposed FM-GCN jointly  extracts the context interactions and aggregates  spatial influences, thus the graph convolution structure enjoys the flexible kernel numbers to further generate feature interactions,  and ultimately squeeze them to $N$-dimension vector which represents region-wise uncertainty.

	\subsubsection{Internal and external uncertainty aggregation}
	We have described the period-level internal content uncertainty and spatiotemporal external context uncertainty in previous sections, where ${\bm{U}_{I}^{(m,i)}}$ and ${\bm{U}_{E}^{(m,i)}}$ are elements in two respective tensors ${\bm{U}_{I}}$ and ${\bm{U}_{E}}$, referring to two sources of uncertainties in region $i$ at period $m$. In this section, to jointly optimize these two kinds of uncertainties and incorporate temporal uncertainty  evolutions, we propose a period-wise LSTM. We realize this process in two steps, 1) combining the two kinds of uncertainties into period-wise overall uncertainty $\bm{U}_o^m$ with an aggregation function $\bm{{\rm{aggr}}(\cdot)}$ , 2) capturing  temporal evolution patterns of uncertainties among different periods with Content-Context uncertainty LSTM (C2-LSTM). The citywide unified uncertainty of the $T+1$ interval ${\bm{U}^{T + 1}}$ can be predicted by,
	\begin{equation}
	\bm{U}_o^m = \bm{{\rm{{aggr}}}}(\bm{U}_I^m,\bm{U}_{E}^m)
	\end{equation}	
	\begin{equation}
	{\bm{U}^{T + 1}} = \bm{{\rm{{C2-LSTM}}}}(\bm{U}_o^m ;{\bm{\theta} _{CL}}{\rm{)}}
	\end{equation}
	where $\bm{\theta} _{CL}$ denotes parameters in the C2-LSTM,  and $\bm{{\rm{aggr}}(\cdot)}$   can be instantiated as concatenation or matrix-based fusion. 
	
	\subsubsection{Active hierarchical uncertainty learning}
	Epistemic uncertainty can be explained as distribution differences between the samples input and samples have been trained, while aleatoric uncertainty is explained as the inherent noise and random influences that cannot be explained explicitly. Based on that, instead of passively learning uncertainties with multiple times of training, we design a hierarchical data turbulence scheme to imitate the OOD samples, slightly noisy samples for actively learning epistemic and aleatoric uncertainty. This data turbulence mechanism is implemented through adding different degrees of Gaussian noises into existing samples with a noise injection function $\textbf{Noise}(\cdot)$. The noise injections can perform drastic turbulence and  tiny drift to simulate OOD and noisy samples, respectively. For a specific region $r_i$ at the interval $\Delta t$,  the corrupted observation value $H_{{C_i}}^{\Delta t}$ can be derived by,
	\begin{equation}
	H_{{C_i}}^{\Delta t} = {\textbf{Noise}}(H_i^{\Delta t})  
	\end{equation}
	
	Even though, the core challenge of lacking definite labels in uncertainty quantification still remains unresolved. To guide the uncertainty learning, we regard this task as weak supervised learning and subsequently propose two indicators of both data quality and unified spatiotemporal uncertainty, then we can actively quantify the uncertainty in cooperation with above data turbulence scheme. 
	
	Firstly, based on above analysis of the active hierarchical sample turbulence, our neural data-quality estimation is analogous to epistemic uncertainty modeling, thus data-quality estimation can not only serve to quantify random noise in aleatoric perspective, but has another role of detecting the OOD samples in epistemic perspective. The quantified degree of region-wise noise is directly measured by the absolute error between original samples and noisy samples, and consequently can  be viewed as the weak supervised information in our data quality estimation. 
	Formally, we present the  weak supervised data quality indicator in interval $\Delta t$ as,
	\begin{equation}
	\sigma _{qua}^{(i,\Delta t)} = |H_i^{\Delta t} - H_{{C_i}}^{\Delta t}|
	\end{equation}
	For period-level uncertainty quantification, we also need to average the interval-level data quality into period-level, e.g $\sigma _{qua}^{(i,m)}$ at the $m$-th period. 
	
	Secondly, to achieve final predicted uncertainty, we need to find an informative indicator of unified period-level uncertainty. Here we refer to variance, which is a statistic for  dispersion measurement and can be seen as the potential variations and uncertainty in regression tasks~\cite{postels2019sampling, hafner2020noise}. Inspired by the spatial proximity, temporal periodicity and closeness in spatiotemporal data, we hereby propose a spatiotemporal variance  as a weak supervised loss to learn the uncertainty mapping functions from historical observations. We first define ${\rm stdv}({\mathcal{V}^*})$ as the function for computing the standard deviation for a set of values in $\mathcal{V}^*$. Specifically, for region $r_i$ at period $m$, the proposed spatiotemporal variance is determined by three views. \textbf{1) Spatial view:}  Associated with  the  adjacent matrix $\bm{A}^t$, we select a set of neighboring observations of $r_i$ and calculate the standard deviation of the observations for each interval\footnote{Here we select the top-5\% most spatially nearest  regions as its neighbors.}, and take the average value of the deviations of all intervals into period-level as ${var}_{s}^{(m,i)}$. \textbf{2) Inter-period view:} We retrieve observations of the same intervals in each  $q+2$ period, measure the  observation deviations for these intervals, and compress these deviations into a  period-level variance ${var}_{\mathit{ep}}^{(\cdot,i)}$. \textbf{3) Intra-period view:} We calculate the interval-wise standard deviations of observations for each period as the intra-period variance ${var}_{\mathit{ip}}^{(m,i)}$. By denoting the observation in $j$-th interval of period $m$ at region $r_i$ as $ \widetilde{H_{{r_i}}^j}(m)$, we formally have, 	
	\begin{equation}
	{\mathop{var}} _S^{(m,i)} = \frac{1}{p}\sum\limits_{j = 1}^p {\mathop {{\rm stdv}}\limits_{{r_k} \in \mathcal{N}({r_i})} (\widetilde{H_{{r_k}}^j}(m))} 
	\end{equation}
	\begin{equation}
	{\mathop{var}} _{ep}^{(\cdot,i)} = \frac{1}{p}\sum\limits_{j = 1}^p {\mathop {{\rm stdv}}\limits_{b \in \{ 0,...,q + 1\} } (\widetilde{H_{{r_i}}^j}(b))} 
	\end{equation}
	\begin{equation}
	{\mathop{var}} _{ip}^{(m,i)} = \mathop {{\rm stdv}}\limits_{j \in \{ 1,2,..,p\} } (\widetilde{H_{{r_i}}^j}(m))
	\end{equation}		
	where $\mathcal{N}(r_i)$ is  the neighboring region set of $r_i$. For simplicity, we average the three types of variances as the spatiotemporal variance indicator ${var}_{ST}$ in the specific spatiotemporal domain, which is written as,
	\begin{equation}
	{{{var}_{ST}}^{(m,i)}} = {\rm Avg}({\mathop{var}} _S^{(m,i)},{\mathop{var}} _{ep}^{(m.i)},{\mathop{var}} _{ip}^{(m,i)})
	\end{equation}
	where ${\rm Avg}$ is the average aggregation function. Then these data quality and spatiotemporal uncertainty indicators will correspondingly change with the data turbulences for uncertainty quantification.
	

	\subsection{Gated Mobility-uncertainty Re-calibration bridge}
	The objective of our uncertainty quantification can be generally summarized as two aspects, to learn what the model does not know, and to maximumly boost the prediction performance. Hence, it is of great significance to further capture the  reciprocity and  interactions between the uncertainty and predicted results. We argue that uncertainties can be decomposed into irreducible variation which can be seen as the inherent randomness, and the complementary parts to prediction results, which may be reducible and helpful to prediction task. Based on this intuition, we propose a Gated Mobility-uncertainty Re-calibration (GMuR) bridge to proactively learn the complementary parts and interactions between point estimation and uncertainty quantification, cooperatively benefiting both tasks from each other. The idea of GMuR is to learn how uncertainty variations impact prediction results and subsequently reduce the uncertainty itself. Formally, we denote the Gate as $f_{gate}$, then the calibrated mobility intensity and uncertainty can be written as:
	
	\begin{equation}
	{f_{gate}} = \tanh ({\bm{W}_{gate}}(\widehat{{\bm{U}^{T + 1}}} \circ \widehat{{\bm{H}^{T + 1}}}))
	\end{equation}
	
	\begin{equation}
	\widehat{{\bm{H}^{T + 1}}} = \widehat{{\bm{H}^{T + 1}}} + \widehat{{\bm{U}^{T + 1}}} \cdot {f_{gate}}
	\end{equation}
	
	\begin{equation}
	\widehat{{\bm{\sigma} ^{T + 1}}} = \widehat{{\bm{U}^{T + 1}}} - \widehat{{\bm{U}^{T + 1}}} \cdot {f_{gate}}
	\end{equation}
	where $\circ$ is the concatenation operation, and  $\bm{W}_{gate}  \in {\mathbb{R}^{N \times 2N}}$ is the learnable weight to map the concatenation of learned uncertainty and predicted results to an $N$-dimension gate. Here we select  {\rm tanh} as the activation function to allow both positive and negative variations to transfer into predictions.
	
	\subsection{Loss function and optimization}
\begin{figure}[!ht]	
		\centering
		\includegraphics[scale=0.30]{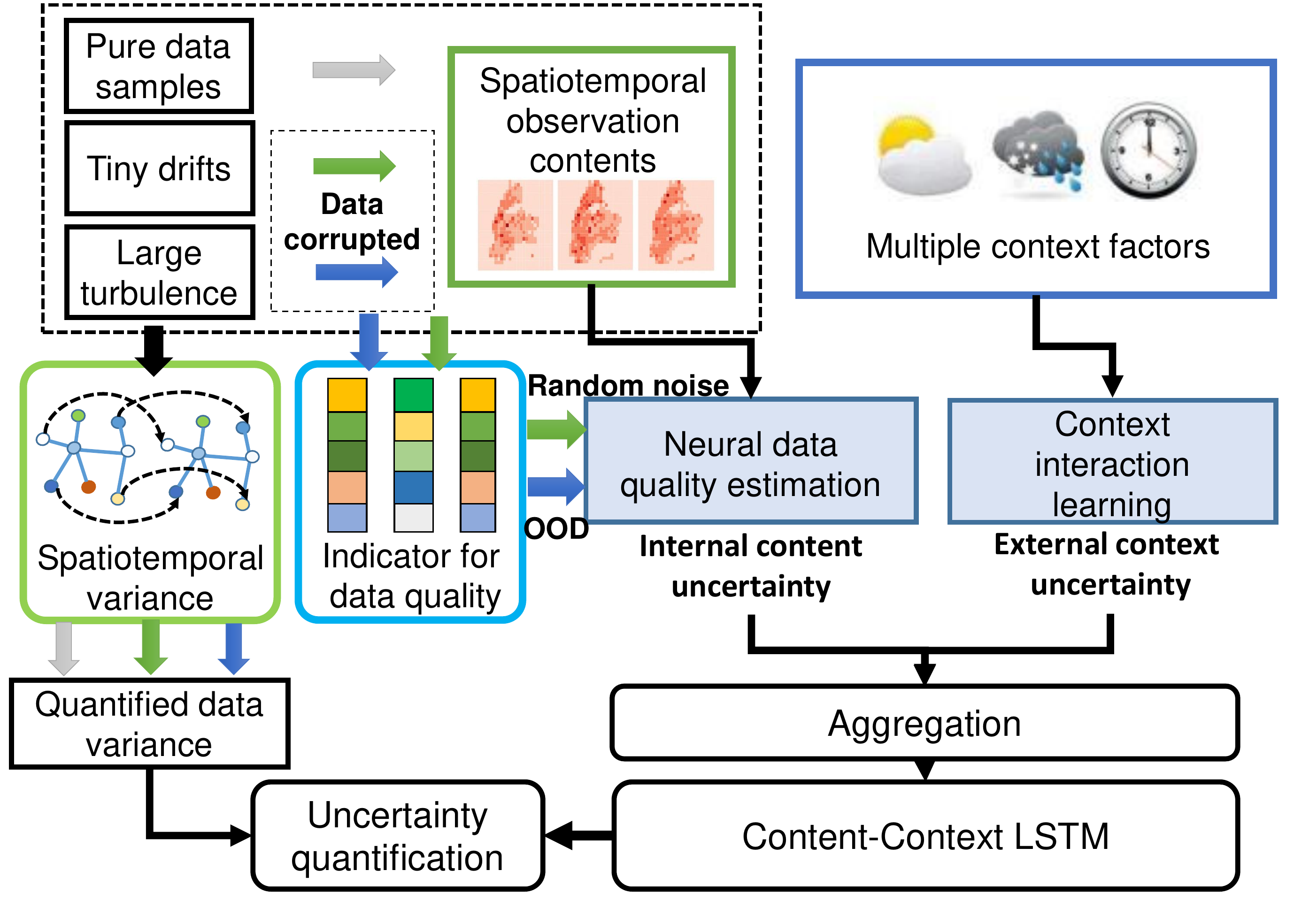}
		\caption{Illustration of active and hierarchical uncertainty learning in C2UQ}
		\label{fig:C2UQ}
	\end{figure}

	With two weak indicators, GMuR bridge, and the hierarchical data turbulence, we can finally perform the active hierarchical uncertainty quantification, as it is briefly illustrated in Figure~\ref{fig:C2UQ}. To take advantage of these uncertainty indicators, we accommodate the data quality and spatiotemporal variance for period-wise guidance, and followed by a last-interval forecasting. 	Thus, we are expected to minimize the following loss in Eq~\ref{eq:loss}, in which we capture  uncertainty evolution on aggregated period-levels with the former two terms, as  well as predict mobility intensity and quantify uncertain fluctuations in last interval with the latter two terms.
	\begin{equation}
	\small
	\begin{split}
	Loss(\Theta ) = \sum\limits_{m = 0}^{q + 1} {\sum\limits_{i = 0}^{N-1} {({{(U_I^{(m,i)} - \sigma _{qua}^{(m,i)})}^2}  + {{(U_o^{(m,i)} - {\mathop{var}} _{ST}^{(m,i)})}^2}) + } } \\ \sum\limits_{i = 0}^{N-1} {({{(\widehat{{{\sigma}_{i} ^{T + 1}}} - {\mathop{var}} _{ST}^{(T + 1,i)})}^2} + {{(\widehat{H_i^{T + 1}} - H_i^{T + 1})}^2})} +  {\left\| {\bm{\sigma} ^{T + 1}} \right\|^2}	
	\end{split}
	\label{eq:loss}
	\end{equation}
	where $\Theta$ denotes  the set of learnable parameters including all $\bm{\theta}_*$ and $\bm{W}_*$, and ${\left\| \cdot  \right\|^2}$ denotes L2-norm for regularizing uncertainties from explosion.  The item $\sigma_{qua}^{(m,i)}$ is enabled when data turbulence is utilized, and we  will directly learn the inherent randomness in existing observations if turbulence is not utilized. For optimizing the algorithm, we introduce Adam optimizer to train our STUaNet~\cite{kingma2014adam}. 
	
	\section{Empirical studies}
	
	\subsection{Data description}
	We use three real-world datasets to verify the effectiveness of uncertainty quantification, and the statistics of these datasets are listed in Table~\ref{tab:dataset}.
	
	\textbf{NYC Taxi.} This dataset consists of approximate 7.5 million taxicab trip records including both pick-up and drop-off events from Jan 1st 2017 to May 31th 2017 in online ride-hailing services. It can typically be an indicator of human mobility where pick-ups and drop-offs stand for departures and arrivals in a specific region\footnote{Here we utilize pick-up events for evaluation.}. 
	
	\textbf{SIP Surveillance.} This dataset contains traffic volumes at 108 interactions in intelligent transportation system, covering the urban area of 45.5 ${km}^2$ in Suzhou Industrial Park. We here utilize dataset from Jan 1st 2017 to March 31st 2017.
	
	\textbf{California Check-ins in Gowalla.} It is a widely used Location-Based Social Network dataset, and contains a total of 736 k check-in records over the period from Feb 1st, 2009 to Oct 31st, 2010~\footnote{ http://snap.stanford.edu/data/loc-gowalla.html}. We choose the state-wide check-ins of California (Carlifo), by filtering longitude and latitude, and then cluster these POIs into 1,200 disjointed regions.  
	
	{Even though the predictions and uncertainty quantification are correlated with the  spatiotemporal scales, the solution evaluations are orthogonal to the generalities of our proposals, based on common urban division settings and fair comparison mechanisms~\cite{zhang2017deep}. For simplicity, we fix the time interval as 30 minutes except 1 hour for California check-ins.}
	\begin{table}[]
		\centering
		\caption{Dataset  statistics}
		\small
		\begin{tabular}{ccccc}
			\toprule
			\hline
			Dataset               & \tabincell{c}{Categroy \\ of datasets}           &  \tabincell{c}{ \# of \\ records}                            & \tabincell{c}{Granularity of \\ region division}             &  \tabincell{c}{\# of \\ regions }          \\ \hline
			NYC   & Taxi  trips                             & 7.5 million & 1.5 $\times$1.5 $km^{2}$  & 354            \\
			SIP  & Surveillance                        & 2.7 million & 0.5 $\times$ 0.5 $km^{2}$ & 108                \\ 
			Carlifo       &Check-ins                          & 	736 k  & Cluster-based	  &   1200              		\\
			\hline
			\bottomrule
			\label{tab:dataset}
		\end{tabular}
	\end{table}

	\subsection{Performance Comparison}
	\subsubsection{Implementation details} 
	We organize our datasets into training samples and divide the samples into 60\%, 30\% and 10\% for training, testing and validation. The initial learning rate is set to 0.001 with an 0.98 attenuation rate every 10 epochs. All methods are implemented in Tensorflow 1.15.0 and trained with 2 Tesla v100 GPUs. We stack 2 GCN layers and 2 LSTM layers in each spatial and sequential learning block, and instantiate $\rho = 0.6, p = 6, q = 3$\footnote{All these hyperparameters are set according to references~\cite{bai2019stg2seq,zhang2017deep} and also fine-tuned carefully, we omit the process due to limited space in this paper.}.
	\subsubsection{Evaluation metrics}
	Given the predicted point estimation ${\widehat{H_i^t}}$, uncertainty quantification ${\widehat{\sigma_i^t}}$, and ground truth ${H_i^t}$ at the region $i$ during interval $t$, we evaluate the effectiveness of our model from aspects of both prediction accuracy and uncertainty quantification quality. 
	Regarding prediction accuracy, we employ RMSE and MAPE for evaluation.
	To evaluate uncertainty learning quality and verify whether the predicted intervals considering uncertainty can accurately capture the ground truth, we introduce the the prediction interval coverage probability (PICP) metric according to~\cite{wang2019deep}, which is defined as 
	\begin{equation}
	{\mathit {PICP}} = \frac{{{C_{obj}}}}{{NT}}
	\end{equation}
	\begin {equation}
	{C_{obj}} = \sum\limits_{t = 0}^{T - 1} {\sum\limits_{i = 1}^N {{\rm{II}}(\widehat{H_i^t} - \widehat{\sigma _i^t} < H_i^t < \widehat{H_i^t} + \widehat{\sigma _i^t})} } 
\end{equation}
where ${{\rm{II(\cdot)}}}$ is an indicator function.

\subsubsection{Effectiveness of collective human mobility forecasting}
In terms of forecasting tasks, we compare our STUaNet against some representative spatiotemporal prediction methods.
\textbf{(1) STG2Seq:} It uses a hierarchical graph convolution model to capture both spatial and temporal dependencies for passenger demand forecasting.
\textbf{(2) STGCN:} This work designs a novel complete convolution structure for comprehensive spatiotemporal correlation modeling in human mobility.
\textbf{(3) MDL:} It is a state-of-the-art collective human mobility forecasting method which inherited from ST-ResNet and simultaneously models nodes and edges  with multiple deep learning tasks.

The upper half of Table~\ref{tab:compar-performance} illustrates the forecasting comparison results. By carefully considering uncertainty quantification and gravity model-based mobility transitions, our method can consistently  outperform three baselines on all datasets. More excitingly, STUaNet surpasses the best baseline DCRNN, STGCN, STG2Seq 13.11\%, 26.59\% and 47.21\% on the metric of MAPE in SIP, NYC and Califor, respectively. We also replace the dynamic adjacent matrix with a static distance-based matrix in our STUaNet for an ablative evaluation, and the performance decreases on STUaNet-Static $\bm{A}$ can prove the necessity of mobility transitions. These promising results  on all three datasets verify that our uncertainty quantification is solid based on this forecasting framework. 

\subsubsection{Effectiveness of uncertainty quantification}
Next, we evaluate the capacities of uncertainty estimation and prediction calibration in different uncertainty learning baselines. We here employ four popular uncertainty quantification mechanisms as baselines.
\textbf{(1) NLL loss:}	The negative log likelihood (NLL) loss  is utilized  to perform station-level Numerical Weather Forecasting (NWF) and the associated uncertainty quantification~\cite{wang2019deep}. 
\textbf{(2) Dropout-based BNN:} We realize this BNN method with dropout~\cite{gal2016dropout, gal2017concrete}, and this mechanism is widely applied in uncertainty quantification for numerous risk-sensitive tasks, ranging from computer vision ~\cite{kendall2017uncertainties,leibig2017leveraging} to NWF~\cite{vandal2018quantifying,liu2020probabilistic}.
\textbf{(3) DeepEnsembles:} We perform the uncertainty learning with the ensemble method which trains a series of neural networks with different initializations~\cite{lakshminarayanan2017simple}\footnote{The number of ensembled networks is set as 5, according to~\cite{lakshminarayanan2017simple}. }. 
\textbf{(4) SDE method:} This is a state-of-the-art uncertainty learning model with injections of noise  and OOD samples, and we reproduce this method by referring~\cite{kong2020sde}. 

All numerical results on uncertainty quantification  are reported in the bottom half of Table~\ref{tab:compar-performance}. Overall, the proposed STUaNet with C2UQ achieves best performance on almost all metrics over three datasets regarding both forecasting and uncertainty learning.  Integrated with C2QU, STUaNet improves the PICP in SIP from 68.83\% of DeepEnsembles to 80.74\%, increasing 17.30\%, and can also obtain comparable accuracy with DeepEnsembles in both NYC and Califor. The slight decrease of PICP in NYC may be attributed to its imbalanced distribution of mobility. In addition, almost all uncertainty-aware forecasting can perform better than non-uncertainty-equipped methods on MAPE metric, which demonstrates the necessity and superiority of spatiotemporal uncertainty quantification. And the higher prediction accuracy of our C2UQ can boil down to GMuR bridge with prediction re-calibrations. 

For a detailed analysis, SDE surpasses all other baselines on forecasting metrics. Its core idea can be viewed as a denoise autoencoder mechanism where noisy and pure observations are trained alternately. The superior results illustrate the effectiveness of alternate training between in-distribution, noisy and OOD samples. However, it has a relatively lower PICP due to lacking  uncertainty labels  for exact quantified uncertainty learning. The uncertainty quantification with only negative log likelihood performs worse than other methods on predictions and shows an instable learning process. This further provides evidence for the intuition of separating learning uncertainty and prediction values. The relatively higher PICPs of NLL and DeepEnsembles are mostly because NLL and ensembles usually derive a larger uncertainty without any guidance.	
Monte-Carlo Dropout-based BNN and DeepEnsembles illustrate a stable training process and can achieve favorable performances, which benefits from the ensembled mechanisms where statistical moment estimations are employed for uncertainty quantification. 

In summary, we argue that these methods are less effective than ours on uncertainty learning in two aspects. First, they are not tailored for spatiotemporal modeling, whichs fail to extract spatiotemporal evolutions and content-context interactions.  Second, ensembles usually require multiple times of training which cost much memory and computation while our C2QU enjoys the efficiency of one-time training. Finally, we also integrate the well-performed STG2Seq with our C2UQ for uncertainty quantification and the results demonstrate the scalability and generality of C2UQ.	
\begin{table*}[]
	\small
	\centering
	\caption{Performance comparisons on three datasets}
	\label{tab:compar-performance}
	\begin{tabular}{c|c|ccc|ccc|ccc}
		\toprule		
		\hline						
		&\multirow{2}{*}{Methods}          &\multicolumn{3}{c|}{SIP}    & \multicolumn{3}{c|}{NYC}             & \multicolumn{3}{c}{California} \\ \cline{3-11}
		&                   & RMSE   & \tabincell{c}{MAPE } & \tabincell{c}{PICP } & RMSE   &  \tabincell{c}{MAPE } &\tabincell{c}{PICP } & RMSE         & \tabincell{c}{MAPE } &  \tabincell{c}{PICP } \\ \hline
		\multirow{4}{*}{\tabincell{c}{Baseline for \\ spatiotemporal \\ prediction} } 
		& STG2Seq           & 3.905  & 31.66\%      & -          & 2.623  & 17.61\%       & -          & 2.545       &30.50\%      & -          \\
		& STGCN             & 4.509  & 35.84\%      & -          & 2.102  & 16.47\%       & -          & 2.621      &32.48\%       & -          \\
		& MDL               & 2.967  & 34.45\%      & -          & 2.984  & 18.62\%       & -          & 1.917      &35.39\%       & -          \\  
		& DCRNN               & 4.750  & 20.21\%      & -          & 3.185  & 37.45\%       & -          & 4.120      &40.02\%       & -          \\  
		
		& STUaNet-Static$\bm{A}$    & 3.358  & 19.35\%      & -          & 1.644  & 12.90\%       & -          & 2.276      &22.36\%       & -        \\  		\hline
		\multirow{4}{*}{\tabincell{c}{Baseline for \\ uncertainty \\ learning} }    
		& NLL               & 5.040   & 48.76\%      & 66.53\%      & 1.930   & 27.83\%   & 70.17\%   & 2.719     & 74.19\%      & 72.90\%      \\
		& DeepEnsembles & 4.299  & 29.36\%      & 68.83\%      & 1.247  & 23.84\%      & 74.25\%      & 2.667      &30.98\%   & 42.66\%      \\ 
		& Dropout-based BNN & 4.086 & 22.97\%      & 61.86\%      & 2.697 & 13.09\%      & 74.32\%   & 2.668     & 26.78\%      & 67.50\%      \\
		& SDE               & 3.086 & 20.32\%      & 60.96\%      & 2.065 & 11.87\%       & 67.97\%   & 2.332     & 18.27\%      & 87.67\%      \\  \hline
		\multicolumn{2}{c|}{\textbf{STUaNet(Ours)}}                        & \textbf{2.942}  & \textbf{17.56\%}      & \textbf{80.74\%}      & \textbf{1.624} & \textbf{11.79\%}     & \textbf{72.14\%}          & \textbf{2.586}       & \textbf{16.10\%}       & \textbf{88.46\%}          \\ \hline
		\multicolumn{2}{c|}{STG2Seq+C2UQ}                 &      3.528      & 22.41\%         &  78.64\%         &  2.320      &  12.96\%   &  70.12\%      &     2.548     &   22.52\%        &  78.80\%    \\\hline
		\bottomrule
	\end{tabular}
\end{table*}




\subsubsection{Quality of uncertainty learning in different intervals.}
\begin{figure}[!ht]	
	\centering
	\includegraphics[scale=0.22]{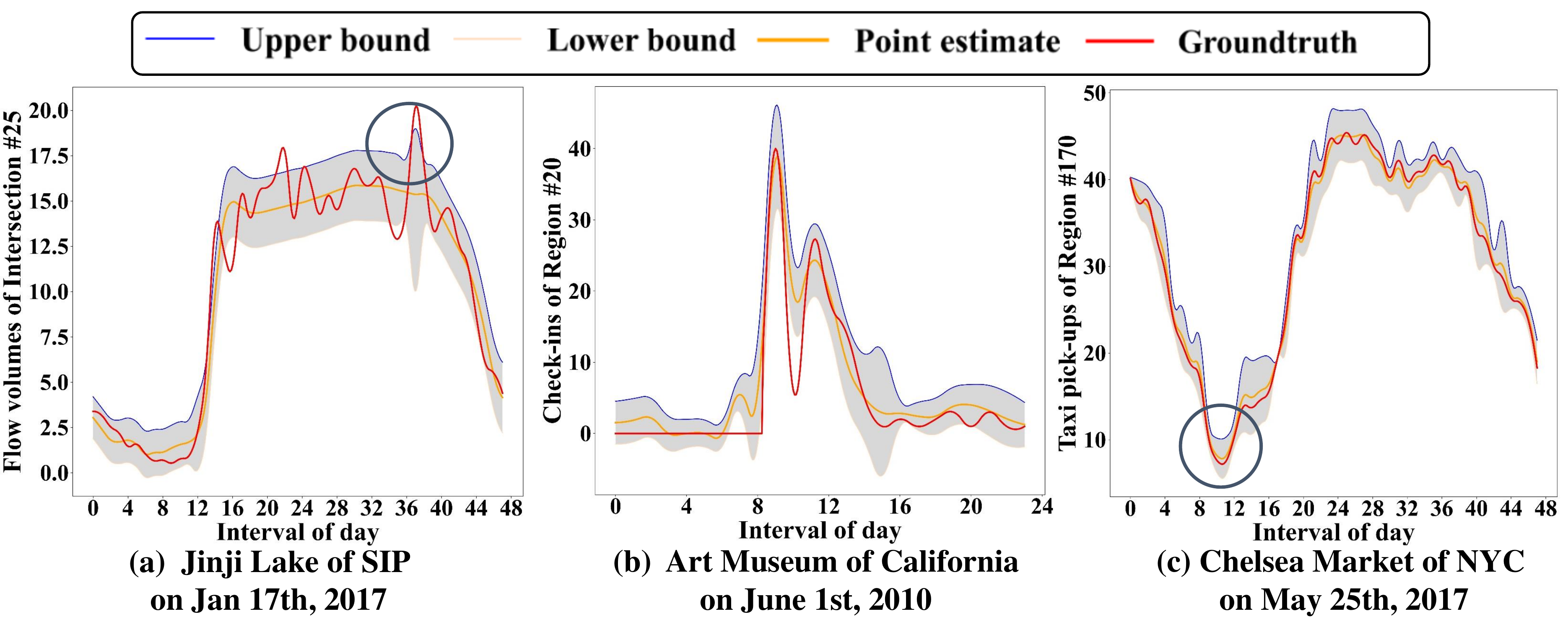}
	\caption{Quality visualization of uncertainty quantification in different datasets}
	\label{fig:quality of uncertainty}
\end{figure}

To provide an intuitive visualization of our uncertainty quantification quality, we choose one typical region in each dataset to illustrate the interval-level prediction and uncertainty results in Figure~\ref{fig:quality of uncertainty}. 
By focusing on the microscopic perspective of our results, we find that the predicted uncertainties are mostly consistent with the prediction errors.  As observed, the prediction interval doesn't become wider or fluctuate heavily over time, and instead, it presents that the widths during nights are mostly narrower than daylights. The main reason lies in that the mobilities are more stable at nights. The circled inaccurate predictions and large uncertainty fluctuations are evening and morning peak hours in SIP and NYC where both suffer rains. We can infer that the concurrent contextual scenarios like rains   occur less frequently in training sets and thus increase the uncertainty on out-of-distribution observations. Thanks to uncertainty prediction, these uncertainties in predicted results can be effectively exhibited for a more reliable decision, thus citizens and administrations can prepare well for the possible uncertainty conditions beforehand. Hence, our uncertainty-aware spatiotemporal forecasting  can provide  more informative and valuable quantified decision-making basis for urban trip planning and city safety.

\begin{figure*}[!]
	\centering
	\includegraphics[scale=0.46]{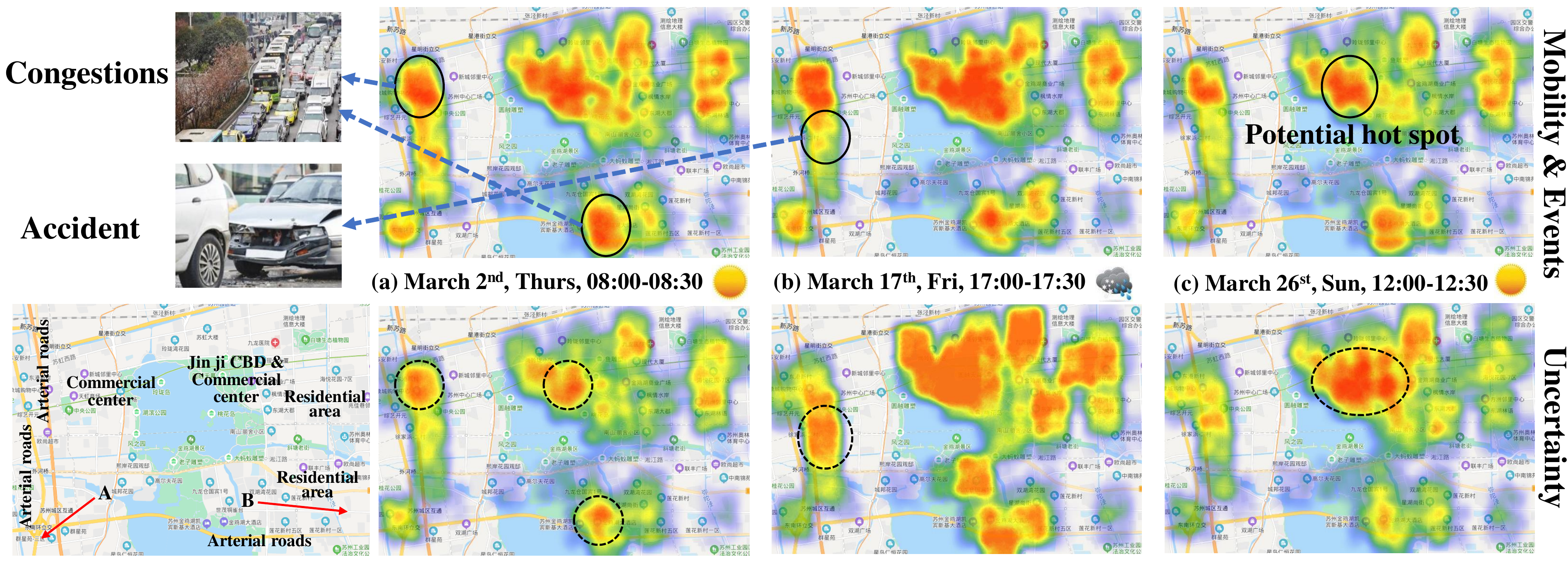}
	\caption{Uncertainty studies in SIP. Solid circles notate the ground-truth urban events while dashed circles highlight regions with high uncertainties for comparison. }
	\label{fig:case-study}
\end{figure*}
\subsection{Ablation study}
We conduct ablation studies to test the sensitivity of each component in our integrated STUaNet. We successively remove typical components as variants of C2UQ. \textbf{C2UQ-1:} Remove the neural data-quality estimation. \textbf{C2UQ-2:} Omit the FM-GCN  and spatiotemporal variance components. \textbf{C2UQ-3:} Expurgate the GMuR bridge. \textbf{C2UQ-4:} Omit the alternate training process and only train on pure dataset.

Table~\ref{tab:ablation-study} illustrates the performance of four ablative variants on three datasets. From the quantitative results, we can see the integrated C2UQ outperforms all its variants. In particular, SIP and NYC datasets are more sensitive to data quality estimation and GMuR modules, while performances on Califor dataset are largely improved by alternate training process. After removing the FM-GCN module, it becomes difficult to capture uncertainty without considering the interactions of context factors and variance-based weak supervised information, by illustrating a prominent decreased performance on three datasets. In contrast, our C2UQ can encourage larger uncertainty for higher spatiotemporal variance and vice versa, where we can actively learn uncertainties. The decreased performance of C2UQ-3 and C2UQ-4 demonstrate the success of re-calibrating spatiotemporal predictions with uncertainty-aware mechanism, and eventually facilitates uncertainty quantification tasks with weak supervised learning. 
\begin{table}[]
	\centering
	\small
	\caption{Performances of ablative variants on three datasets}
	\label{tab:ablation-study}
	
	\begin{tabular}{c|c|c|c}
		\toprule
		\hline
		\multirow{2}{*}{Variants}&  SIP & NYC & Califor     \\ \cline{2-4}
		&  {MAPE/PICP} &{MAPE/PICP} & {MAPE/PICP}     \\ \hline
		C2UQ-1       & 33.13\%/73.35\%   & 28.70\%/58.07\% & 9.24\%/72.40\%   \\
		C2UQ-2       & 21.29\%/70.21\%  & 27.96\%/60.59\% & 14.39\%/69.31\%  \\
		C2UQ-3       & 28.20\%/63.44\%   & 10.65\%/63.39\%   & 22.93\%/67.43\%  \\
		C2UQ-4       & 24.86\%/80.80\% & 36.20\%/71.50\%   & 20.78\%/67.00\%  \\
		\textbf{C2UQ}& \textbf{17.56\%/80.74\%}   & \textbf{11.79\%/72.14\%} &\textbf{16.10\%/88.46\%}	\\\hline
		\bottomrule
	\end{tabular}
	
\end{table}

%

\begin{figure*}[!ht]	
	\centering
	\includegraphics[scale=0.55]{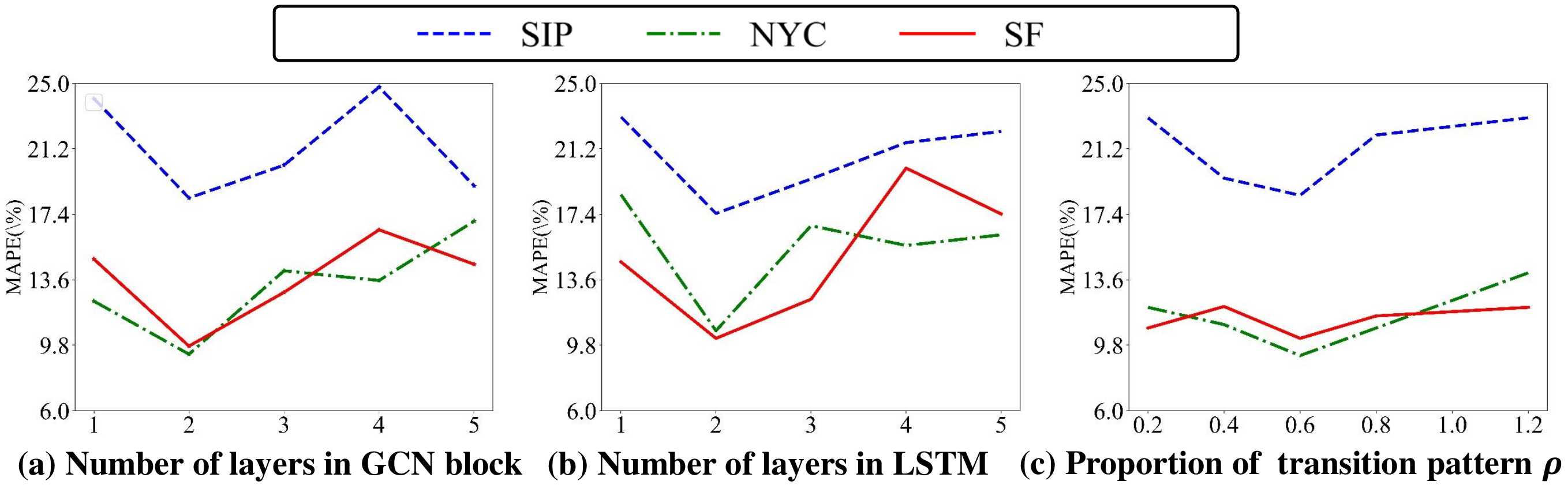}
	\caption{Performance on different parameter settings}
	\label{fig:Hyper-study}
\end{figure*}
\subsection{Case study}
A higher uncertainty indicates that the prediction model is not confident about the predicted value or there exists a large dispersion among its historical observations under this context. In this section, we generate the mobility and uncertainty maps with STUaNet from test sets, to investigate how can they benefit diverse web applications.

\textbf{(1) Urban event detection and prediction.} Figure~\ref{fig:case-study}(a) illustrates the urban situation of peak morning hours on March 2nd. 
Under the context of sunny morning, these three highlighted  regions with both high uncertainties and mobility intensities can be interpreted as urban events like congestions, which are further verified in ground truth. These predictions motivate travelers to re-plan their routes and urge traffic agency to proactively evacuate crowds to avoid urban safety concerns like accidents and spread of pandemics. With a heavy rain, the increasing uncertainties across urban regions in subfigure (b) reveal our model lacking the confidence in such prediction, due to the rare weather event and complicated context interactions. This not only reflects the principle of epistemic uncertainty, also verifies the common practice that the increasing probability of burstiness like accidents on rainy days can reasonably contribute to mobility uncertainties. Therefore, it is of great significance to provide uncertainty-aware predictions which actively prevent misleading decision-makings. \textbf{(2) Mining potential commercial interests.} As shown in subfigure (c), there exhibits an expansive coverage of higher uncertainty with moderate mobility intensity around Jinji Commercial Center, which implies the potential mobility fluctuations during following intervals. For businessmen, crowds are profits, thus they can take advantage of these uncertainties and preferable weather, to maximumly motivate buying desires of consumers and drive the uncertainty into a positive increase of intensities, by propagandizing the sale promotions with online web applications. \textbf{(3) Deeper understanding the nature of human mobility.} We can also discover several interesting phenomena. Firstly, we identify that commercial centers are more sensitive to weather changes while arterial roads are more stable to context, and particularly the Jinji Circle is mostly with high uncertainties as it may experience the quick and dynamic flow changing for its integratedly complicated functionalities. Secondly, we can also provide urban planning suggestions for regions A and B to build some commercial complex for attracting the mobilities as they are currently with both lower uncertainty and volume intensity. By uncertainty learning, we can dive deeper into human mobility, uncover the potential intentions and facilitate the urban planning and human-centered computing for a better life.

\subsection{Hyperparameter study}
To investigate how different values of hyperparameters  impact the prediction performance, we show the hyperparameter studies 
here. The hyperparameters  are three-fold here, i.e., the number of GCN layers, the number of LSTM layers and $\rho$ in adjacent matrix. We show the fine-tuning process  in Figure~\ref{fig:Hyper-study} and for simplicity, we only compare the metric of MAPE in regression tasks which is more fair and intuitive for different datasets. 
Finally, we stack 2 GCN blocks, 2 LSTM layers, and set  $\rho$ = 0.6 on all three dataset learning tasks. 
%


\section{Conclusion and discussion} 

STUaNet, which internalizes the uncertainties into the model from the perspectives of internal content consistency, external context interactions and temporal evolutions, is a pioneering attempt on spatiotemporal uncertainty quantification in collective human mobility.
In particular, to tackle uncertainty quantification challenge, we transfer it into a weak supervised learning and an active hierarchical uncertainty learning by proposing two implicit but quantifiable uncertainty indicators. Extensive experiments on three mobility-related datasets verify the effectiveness of our proposal.


For uncertainties, regardless of whether they are aleatoric or epistemic and are internal or external, these uncertainties are both data-dependent and model-dependent. Therefore, we burst forth a bold idea that all these uncertainties can be summarized from an epistemic view by considering unpredictable factors and highly complex interactions as high-level knowledge that should be deeply learned and understood from a long-term perspective. In future, we will further explore both the quantified regularities and uncertainties of spatiotemporal data with more basic and theoretical analysis, and hence explicitly optimize spatiotemporal predictions by identifying the sources of deductible uncertainties.

\begin{acks}
	This work is partially supported by Zhejiang Lab's International Talent Fund for Young Professionals, Anhui Science Foundation for Distinguished Young Scholars (No.1908085J24), NSFC (No.62072427, No.61672487, No.61772492), and Jiangsu Natural Science
	Foundation (No.BK20191193).
\end{acks}
\balance
\bibliographystyle{ACM-Reference-Format}
\bibliography{Uncertainty}

%
%
%
%
%
%
%
%

\end{document}